\documentclass{annals-preprint}

\title{Spatiotemporal Recognition of Abdominal Wall Layers During Optical Trocar Entry}

\author{%
Shreeram Athreya\authormark{1,*},
Andrew Melehy\authormark{2},
Carter Prentice\authormark{2},
Michael Robinson\authormark{2},
Corey Arnold\authormark{1,3,4,5},
James Wu\authormark{2},
and William Speier\authormark{3,4}%
}

\articletype{Original Article}
\shortauthors{Athreya et al}
\shorttitle{Abdominal Wall Layer Recognition During Trocar Entry}
\issueinfo{Preprint \textbullet\ October 2026}
\footertext{}
\publicationstatus{}

\articlenotes{%
\authormark{1} Department of Electrical and Computer Engineering, University of California, Los Angeles, Los Angeles, California, USA.

\authormark{2} Department of Surgery, University of California, Los Angeles, Los Angeles, California, USA.

\authormark{3} Department of Radiological Sciences, University of California, Los Angeles, Los Angeles, California, USA.

}{%
\authormark{4} Department of Bioengineering, University of California, Los Angeles, Los Angeles, California, USA.

\authormark{5} Department of Pathology and Laboratory Medicine, University of California, Los Angeles, Los Angeles, California, USA.

\authormark{*} Corresponding author: Shreeram Athreya, Department of Electrical and Computer Engineering, University of California, Los Angeles, Los Angeles, CA, USA. Email: shreeram@ucla.edu
}

\begin{document}
\maketitle

\begin{abstract}
\abstractitem{Objective}{To evaluate whether spatial context and temporal modeling improve tissue recognition during optical trocar entry, where multiple abdominal-wall tissues may be visible simultaneously.}
\abstractitem{Methods}{We retrospectively analyzed 71 recordings containing 1,516 annotated keyframes and 29,151 clinician-annotated sectors. A polar grid divided each frame into 28 sectors assigned to nine tissue classes. Compact color descriptors, neighboring-sector summaries, and positional features supplied a gradient-boosted emission classifier. The complete pipeline added learned temporal smoothing, Markov random field regularization, an anatomically ordered hidden Markov model, and posterior blending. Five-fold cross-validation was grouped by recording, with cross-fitting used to train the temporal stage without within-recording leakage. Primary evaluation used clinician-derived targets; secondary analyses included transferred labels. Ablations assessed classification, probability quality, spatial consistency, and peritoneum-or-deeper onset timing.}
\abstractitem{Results}{Values are fold means $\pm$ standard deviations. The complete pipeline achieved accuracy of $0.646\pm0.037$, macro-F1 of $0.507\pm0.049$, and macro area under the receiver operating characteristic curve of $0.882\pm0.018$, compared with $0.600\pm0.033$, $0.440\pm0.045$, and $0.861\pm0.024$ for emissions alone. The paired macro-F1 improvement was $0.066\pm0.022$, with gains in all five folds. Label changes decreased from 66.2 to 19.5 per sector-minute, and mean fold-specific median onset error decreased from 15.58 to 2.45 seconds. The proportion of frames with a tissue-depth-code span of at least three decreased from 53\% to 26\%, versus 22\% for clinician annotations. Recognition remained limited for preperitoneal fat and bowel. Learned smoothing alone produced lower log-loss than the complete pipeline, and the latter retained a 44\% premature-onset proportion.}
\abstractitem{Conclusions}{Combining appearance, spatial context, and temporal constraints improved retrospective tissue recognition and onset localization. Modest tissue-specific performance, residual timing errors, and reliance on future observations require external validation and prospective evaluation of causal models before clinical guidance can be assessed.}
\end{abstract}

\section{Introduction}
Establishing abdominal access is a consequential step in laparoscopic surgery. Trocar insertion can cause vascular or visceral injury, and adverse-event reports document serious morbidity and fatalities associated with access devices.\citep{Fuller2005Trocarassociated,Champault1996Serious,Schafer2001Trocar} Comparisons of open, closed, and direct-vision techniques have not established a universally superior approach for preventing major injuries, partly because these events are uncommon and comparative studies are underpowered.\citep{Ahmad2019Laparoscopic,Chapelle2015Trocar} This uncertainty motivates investigation of complementary information that could improve understanding of tissue encountered during entry. Optical trocar video provides a visual record from which tissue appearance and progression can be analyzed.

Advances in surgical computer vision have enabled recognition of operative phases, instruments, anatomical structures, and action relationships.\citep{Twinanda2017EndoNet,Nwoye2022Rendezvous,Madani2022Artificial} These developments support the broader objective of extracting clinically meaningful information from surgical video.\citep{MaierHein2017Surgical,Hashimoto2018Artificial} However, recognition during abdominal-wall traversal poses a different problem from assigning an operative phase to an entire frame. Several tissues may occupy the field simultaneously, and identifying their spatial distribution is necessary to characterize the changing view. A useful representation should therefore accommodate local differences while retaining information about neighboring regions and the sequence of observations. Glare, similar coloration across fat compartments, and thin visible layers further complicate recognition, making local appearance insufficient for some clinically relevant distinctions.

Appearance and sequence information also present competing demands. Independent frame predictions can fluctuate, whereas strong anatomical constraints can suppress legitimate transitions or weak evidence for briefly visible tissues. For the present task, accuracy alone cannot distinguish a useful reduction in noise from excessive smoothing. Recognition must also be assessed through class-specific performance, spatial consistency, and the timing of peritoneum-or-deeper onset. Probability quality deserves separate scrutiny because discrimination and confidence calibration describe different properties of a predictive model.\citep{Guo2017On,Mehrtash2020Confidence} These considerations motivate evaluating the contributions and tradeoffs of individual modeling stages.

We investigate a framework that partitions the circular view into 28 sectors and recognizes nine tissue classes using compact color descriptors, neighboring-sector context, and gradient boosting. Learned temporal smoothing is combined with spatial regularization, an anatomically ordered hidden Markov model, and posterior blending. The study includes 71 recordings and 1,516 annotated keyframes. Evaluation uses five folds grouped by recording, with leakage-resistant cross-fitting for the learned temporal stage and clinician-derived targets for the primary analysis. This design assesses generalization between recordings while distinguishing clinician annotations from automatically transferred training labels. The reported temporal pipeline uses future observations and is evaluated retrospectively. Our contributions are as follows:
\begin{itemize}
\setlength\itemsep{2bp}
\setlength\parskip{0pt}
\setlength\parsep{0pt}
\item \textbf{Spatially resolved tissue recognition.} We formulate optical trocar entry as nine-class recognition over 28 sectors, accommodating multiple tissues within each frame and evaluating 29,151 clinician-annotated sectors.
\item \textbf{Integrated spatial and temporal modeling.} We combine compact appearance features, learned posterior smoothing, anatomical constraints, and blending, using cross-fitting to prevent within-recording leakage during temporal-model training.
\item \textbf{Annotation-aware evaluation.} We retain label provenance, distinguish primary clinician-only evaluation from transferred-label scoring, and compare all models under the same video-grouped cross-validation design.
\item \textbf{Comprehensive experimental characterization.} We report component and feature ablations, unsuccessful alternatives, and classification, probability, spatial, and temporal endpoints, demonstrating macro-F1 improvement from 0.440 to 0.507 while documenting remaining tissue-specific and onset-detection errors.
\end{itemize}

\section{Related Work}
\subsection{Entry Safety and Training}
Laparoscopic entry studies document trocar-related morbidity through surveys, observational cohorts, and adverse event reports.\citep{Yuzpe1990Pneumoperitoneum,Jansen1997Complications,Mayol1997Risks,Fuller2005Trocarassociated,Champault1996Serious,Chapron1998Surgical,Schafer2001Trocar} Vascular-injury reports and management reviews emphasize the consequences of access errors.\citep{Nordestgaard1995Major,Larobina2005Complete,Asfour2018Vascular,Shaikh2023Laparoscopic} Technical studies examine insertion procedures,\citep{Vilos2009Three,Theodoropoulou2008Direct,Dunne2011Establishing,Jansen2004Complications} while reviews compare entry strategies and trocar designs.\citep{Molloy2002Laparoscopic,Vilos2007Laparoscopic,Krishnakumar2009Entry,Alkatout2017Complications,Merlin2003Systematic,Jiang2012The,Elnaggar2023Direct} Cochrane syntheses highlight insufficient evidence for superiority on major injury outcomes, motivating complementary approaches to interpreting tissue during access.\citep{Chapelle2015Trocar,Ahmad2019Laparoscopic}

Training research provides a parallel foundation. Simulation, proficiency-based curricula, and automaticity training address technical performance and skill transfer.\citep{Aggarwal2010Training,Fried2004Proving,Vassiliou2010FLS,Gallagher2005Virtual,Seymour2002Virtual,Grantcharov2004Randomized,Ahlberg2007Proficiencybased,Stefanidis2012Simulator} Studies associate operative skill with complications and examine patient outcomes after mastery learning.\citep{Birkmeyer2013Surgical,Zendejas2011SimulationBased} Objective assessment increasingly uses annotated video and automated analysis,\citep{Vedula2017Objective,Pedrett2023Technical,Gruter2023Videobased,Yanik2022Deep,Loukas2020Surgical,Jin2018Tool} although tissue classification alone cannot establish educational effectiveness or improvements in surgical safety.

\subsection{Surgical Video Recognition}
Surgical data science extends video analysis toward workflow understanding and intraoperative assistance.\citep{MaierHein2017Surgical,Hashimoto2018Artificial,Bodenstedt2020Artificial,Loukas2018Video} Early operating-room state recognition and statistical process models preceded contemporary deep architectures.\citep{Bhatia2007Realtime,Padoy2012Statistical,Lalys2014Surgical} EndoNet combined phase recognition with tool detection; recurrent networks, temporal convolutions, and transformers subsequently expanded sequence modeling.\citep{Twinanda2017EndoNet,Jin2018SVRCNet,Czempiel2020TeCNO,Gao2021TransSVNet} Clinical applications and comparative studies demonstrate broader interest in recognizing procedural context.\citep{Kitaguchi2020Realtime,Garrow2021Machine,Raptis2026Surgical,Mezzina2025Surgeons} These tasks principally characterize operative activities, whereas trocar-entry recognition must distinguish simultaneously visible tissues.

Anatomical segmentation and critical-view assessment connect computer vision more directly to surgical safety.\citep{Madani2022Artificial,Mascagni2022Artificial,Mascagni2021Intraoperative} CholecSeg8k supplies semantic labels; Rendezvous and CholecTriplet2022 address instrument--verb--target relationships.\citep{Hong2020CholecSeg8k,Nwoye2022Rendezvous,Nwoye2023CholecTriplet2022} HeiChole and PhaKIR broaden comparative evaluation across workflow, skills, instruments, and keypoints,\citep{Wagner2023Comparative,Rueckert2025Video,Rueckert2026Comparative} while SurgBench and SurgAtlas expand task and procedural coverage.\citep{Wei2025SurgBench,Bellos2026SurgAtlas} Their respective labels and acquisition settings differ from the abdominal-wall sectors examined here.

\subsection{Temporal Models and Efficient Features}
Structured sequence models offer complementary mechanisms for constraining predictions. Hidden Markov models represent state transitions, semi-Markov models explicitly represent durations, and conditional random fields model conditional label dependencies.\citep{Rabiner1989A,Yu2010Hidden,Lafferty2001Conditional} Temporal convolutional networks learn temporal context and refine action segments.\citep{Lea2017Temporal,Farha2019MSTCN} Our framework combines learned posterior smoothing with explicit spatial and anatomical constraints, without evaluating duration models or these neural architectures.

Efficient visual representations are also relevant to eventual deployment. MobileNet, MobileNetV2, and EfficientNet pursue computational efficiency through architectural design and scaling, while knowledge distillation transfers predictive behavior between models.\citep{Howard2017MobileNets,Sandler2018MobileNetV2,Tan2019Efficientnet,Hinton2015Distilling} These approaches motivate future comparisons with our compact color descriptors; their inclusion does not imply that neural compression or acceleration was evaluated in this study.

\subsection{Annotation and Predictive Reliability}
Reliable evaluation requires attention to annotation provenance, label definitions, and validation design. Surgical annotation studies and consensus recommendations identify these issues,\citep{Pangal2021A,NyangohTimoh2023A,Ward2021Challenges,Eckhoff2023SAGES} alongside work on representative splits and transparent reporting.\citep{Kostiuchik2024Surgical,Tejani2024ChecklistUpdate} Separately, calibration and uncertainty research distinguishes confident predictions from dependable probabilities,\citep{Guo2017On,Mehrtash2020Confidence,Kompa2021Second,Zou2023A,Gawlikowski2023A} including Bayesian approximations and uncertainty decomposition.\citep{Gal2016Dropout,Kendall2017What} Risk-controlling prediction sets, conformal risk control, and learn-then-test procedures provide further frameworks for specifying error targets under their assumptions.\citep{Bates2021Distribution-freeSets,Angelopoulos2024Conformal,Angelopoulos2025LearnControl} These distinctions motivate reporting discrimination, probability quality, and temporal errors separately, while reserving clinical deployment claims for prospective validation.

Within this literature, our study addresses retrospective, spatially resolved nine-class tissue recognition during optical trocar advancement. The comparison examines how appearance, neighboring sectors, and temporal evidence contribute to classification and onset localization. Its contribution is an evaluated integration of established modeling ideas for this setting, rather than a claim that procedural recognition benchmarks or retrospective accuracy demonstrate prevention of vascular or visceral injury during laparoscopic entry.

\section{Methods}
\subsection{Study Design and Video Dataset}
We developed and retrospectively evaluated a spatially resolved framework for tissue recognition during optical trocar entry. The framework assigned a tissue class to each of 28 sectors within the circular laparoscopic field of view (FoV), allowing several tissues to be represented within a single frame. Processing comprised video preparation and alignment, sector-level annotation, color-feature extraction, a frame-independent emission classifier, learned temporal smoothing, spatial regularization, hidden Markov modeling, and posterior blending. Comparisons used five-fold cross-validation grouped by recording.

\begin{figure}[!tbp]
\centering
\includegraphics[width=\columnwidth]{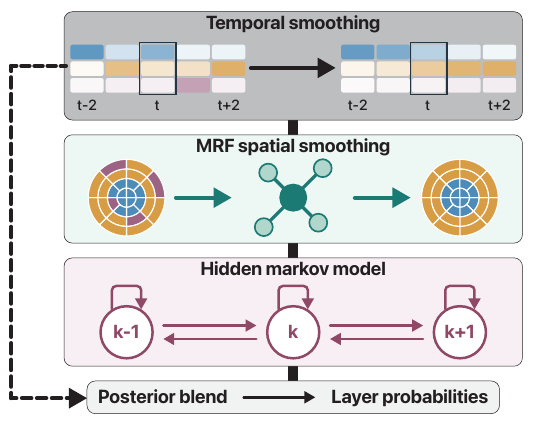}
\caption{Spatial and temporal posterior refinement. Learned temporal smoothing combines emission probabilities across a temporal neighborhood, shown schematically around time $t$. MRF regularization couples neighboring sectors within each frame. The banded HMM permits persistence and transitions between adjacent tissue states $k-1$, $k$, and $k+1$. Posterior blending combines the regularized output with the original emission probabilities, represented by the dashed bypass, to obtain final tissue-class probabilities. The illustrated temporal neighborhood and colors are schematic; the temporal neighborhoods and nine-class vocabulary are specified in the text.}
\label{fig:pipeline}
\end{figure}

The sequence of posterior-processing stages is illustrated in \autoref{fig:pipeline}. Learned temporal smoothing first incorporates neighboring observations, the Markov random field (MRF) regularizes adjacent sectors, and the banded hidden Markov model (HMM) constrains anatomical transitions. Final blending retains evidence from the original emission probabilities.

The recordings had a resolution of $1920\times1080$ pixels. Of the 71 analyzed recordings, 62 were acquired at approximately 60 frames per second (fps; range, 59.99--60.02), and nine at approximately 30 fps (range, 29.97--30.01). Every conversion between frame indices and elapsed time used the individual recording's measured frame rate.

Annotations were created in CVAT on selected keyframes. Masks marked outside the image were ignored, and multiple masks of the same tissue on a frame were combined by logical union. Annotation frame indices were aligned to the corresponding video frames before analysis. The analysis required at least two annotated keyframes per recording. Dataset and evaluation counts are summarized in~\autoref{tab:cohort}.

\begin{table}[!htbp]
\caption{Dataset Assembly and Evaluation Populations}
\label{tab:cohort}
\centering
\footnotesize
\setlength{\tabcolsep}{2bp}
\renewcommand{\arraystretch}{1.10}
\begin{tabular}{@{}p{0.60\columnwidth}p{0.34\columnwidth}@{}}
\toprule
Dataset component & Count or specification \\
\midrule
Analyzed videos & 71 (21 dense; 50 transferred) \\
Approximately 60-fps recordings & 62 (59.99--60.02 fps) \\
Approximately 30-fps recordings & 9 (29.97--30.01 fps) \\
Annotated keyframes with eligible sectors & 1,516 \\
All-label sectors & 36,886 \\
Clinician-only sectors & 29,151 \\
Clinician-only sectors from dense recordings & 9,410 \\
Clinician-only sectors from transferred recordings & 19,741 \\
Observed peritoneum-or-deeper sector onsets & 1,774 \\
Median onset-reference interval width & 4.3 s \\
\bottomrule
\end{tabular}
\par\vspace{4bp}
\raggedright\footnotesize
Primary classification used clinician-only targets; secondary classification used all targets.
\end{table}

\subsection{Tissue Vocabulary and Annotation Paradigms}
Nine classes were assigned integer codes in anatomical progression order: subcutaneous fat, Scarpa's fascia, anterior rectus sheath, rectus muscle, posterior rectus sheath, preperitoneal fat, peritoneum, bowel, and intraperitoneal fat. Background was coded as zero and excluded as a target, as were labels outside this vocabulary. Mesenteric fat was absent from the analyzed data. Codes 7--9 were grouped as peritoneum-or-deeper for onset analyses. Bowel and intraperitoneal fat could alternate in the visible field after entry.

Two annotation paradigms were combined. Under the dense paradigm, the clinician delineated all visible tissue regions on each annotated keyframe. Under the original paradigm, only newly appearing tissues were delineated, and subcutaneous fat was not included in the annotation vocabulary. A deterministic transfer procedure augmented the original annotations with peripheral subcutaneous-fat labels. Training used the resulting combined label set, whereas primary evaluation used targets derived exclusively from clinician-drawn pixels. Thus, annotation provenance was retained through target construction rather than treating automatically filled regions as clinician annotations, addressing a central concern in surgical-video annotation.\citep{Ward2021Challenges,Eckhoff2023SAGES}

\subsection{Deterministic Annotation Transfer}
Transfer was performed independently on each keyframe using the full-resolution FoV. Original clinician masks were retained without boundary regrowth or relabeling. Let $U$ denote their union and $F$ the FoV. Frames satisfying $|U\cap F|/|F|\geq0.65$ were left unchanged. On frames with lower coverage, filling was permitted only when Scarpa's fascia or the anterior rectus sheath was present on that same keyframe. This condition restricted filling to views containing an outer abdominal-wall structure.

Connected components of the unannotated region $F\setminus U$ were identified. Components smaller than 200 pixels were ignored. A component was assigned subcutaneous fat only if it intersected the peripheral annulus $r>0.90R$, where $R$ was the FoV radius. Enclosed unannotated islands, including central openings, remained unlabeled. Subcutaneous fat was the only new tissue class introduced by the adopted procedure. An alternative that filled additional historically plausible tissues was also evaluated.

Transfer was assessed on recordings with both original-paradigm and dense clinician annotations. The dense annotations were used as the comparison reference and were not supplied to the transfer algorithm. Agreement was summarized by mean per-class pixel intersection over union (IoU). For class $c$, this quantity was computed from the intersection and union of transferred and dense-reference pixel sets. Transfer-validation results and the additional-label experiment are reported separately from tissue-classification performance.

\subsection{Video Preparation and Source-Frame Alignment}
Videos were temporally restricted to the annotated span. Each source image was center-cropped to a $1080\times1080$ square and bilinearly resized to $512\times512$ pixels:
\begin{equation}
\begin{aligned}
d&=\min(H,W)=1080,\\
x_0&=\left\lfloor(W-d)/2\right\rfloor=420,\\
I_{512}&=\operatorname{resize}_{512\times512}
  \bigl(I[:,x_0:x_0+d]\bigr).
\end{aligned}
\end{equation}
Color correction was performed after spatial preprocessing. Temporal alignment between annotations and video frames was verified before feature extraction, and only aligned keyframes within the analyzed temporal span were retained.

\subsection{Field-of-View Detection and Polar Geometry}
The circular FoV was estimated from the first full-resolution source frame and retained throughout the recording. The grayscale image underwent Otsu thresholding and morphological opening with a $9\times9$ kernel. The largest connected component was selected, excluding smaller disconnected structures such as captions and logos. For component pixel set $\mathcal{C}$, the fitted center and area-equivalent radius were
\begin{equation}
c_x=\overline{x}_{\mathcal{C}},\qquad
c_y=\overline{y}_{\mathcal{C}},\qquad
R=\sqrt{|\mathcal{C}|/\pi}.
\end{equation}
The circle was mapped into resized-image coordinates by
\begin{equation}
\begin{aligned}
c_x^{(512)}&=(c_x-x_0)512/d,\\
c_y^{(512)}&=c_y512/d,\qquad R^{(512)}=R512/d.
\end{aligned}
\end{equation}

Four concentric rings of equal radial width were used. The innermost ring contained four angular sectors; each of the remaining rings contained eight, giving $(M_0,M_1,M_2,M_3)=(4,8,8,8)$ and 28 sectors in total. The inner annulus had one-seventh the area of the outer annulus, motivating fewer angular subdivisions near the optical axis. For radius $r$ and polar angle $\theta\in[-\pi,\pi]$, sector membership was defined as
\begin{equation}
\begin{aligned}
k&=\min\!\left(\left\lfloor4r/R\right\rfloor,3\right),\\
a&=\left\lfloor\frac{(\theta+\pi)M_k}{2\pi}\right\rfloor\bmod M_k,\\
i&=\sum_{k'<k}M_{k'}+a .
\end{aligned}
\end{equation}
Pixels with $r>R$ were outside the grid. Annotation masks were processed on the $1080\times1080$ crop, while image features were computed at $512\times512$ resolution using the corresponding scaled geometry.

Each sector had inward, outward, left-angular, and right-angular neighbors. Radial neighbors were selected by the sector containing its angular midpoint in the adjacent ring; angular neighbors were indexed modulo the ring's sector count. Where a radial neighbor did not exist, the sector itself supplied the feature-context value. Such self-substitutions were omitted from spatial pairwise interactions.

\subsection{Sector Targets and Primary Evaluation Labels}
On each keyframe, masks were cropped and painted into a class-code map $Y_f$ in XML track order. Later masks overwrote earlier masks where they overlapped. Unannotated pixels retained code zero. For sector pixel set $P_i$, let $A_{f,i}=\{p\in P_i:Y_f(p)>0\}$. A target was assigned only when labeled pixels covered at least one-quarter of the sector:
\begin{equation}
y_{f,i}=\begin{cases}
\displaystyle\arg\max_c\sum_{p\in A_{f,i}}
 \mathbf{1}\{Y_f(p)=c\},&\frac{|A_{f,i}|}{|P_i|}\geq0.25,\\[4pt]
\varnothing,&\text{otherwise}.
\end{cases}
\end{equation}
Unlabeled sectors were excluded as training and evaluation targets but could contribute image features to neighboring-sector context and the frame mean.

For primary evaluation, both the coverage threshold and majority class were recomputed using clinician-drawn pixels only. All labeled pixels were eligible in dense annotations. In transferred recordings, only pixels inside the original masks were eligible, excluding added subcutaneous fat. All primary-evaluation sectors also belonged to the all-label sector set, although the majority target could differ between the two definitions. A single set of held-out predictions was therefore scored against both target sets. Model fitting continued to use all eligible training targets, including transferred labels.

\subsection{Color Features and Specular-Glare Handling}
Features were extracted for every annotated keyframe and every tenth video frame. Their union formed the time grid for temporal inference and full-sequence measurements. At 60 fps, the regularly sampled component had intervals of approximately one-sixth of a second; at 30 fps, the corresponding interval was approximately one-third of a second. Added keyframes introduced intermediate grid locations.

Specular pixels were identified in HSV space from RGB values scaled to $[0,1]$. A pixel was classified as glare when value exceeded 0.92 and saturation was below 0.25. Glare pixels were excluded from histograms and moments, and their proportion within each sector was retained as a separate feature. Each frame underwent independent gray-world white balancing over nonglare FoV pixels. If $\mu_h$ was the mean intensity in channel $h\in\{R,G,B\}$, corrected intensities were
\begin{equation}
\widetilde I_h=\operatorname{clip}\!\left(
 I_h\frac{\frac13\sum_{h'}\mu_{h'}}{\max(\mu_h,10^{-4})},0,1\right).
\end{equation}
This normalization was intended to reduce variation in illumination and color balance and used only the current frame.

Descriptors combined CIE $L^*a^*b^*$ and HSV channels. Conversion to CIE Lab used inverse sRGB gamma correction,
\begin{equation}
g^{-1}(u)=\begin{cases}
((u+0.055)/1.055)^{2.4},&u>0.04045,\\
u/12.92,&u\leq0.04045,
\end{cases}
\end{equation}
followed by the standard linear-RGB-to-XYZ transformation, normalization to the D65 white point $(0.95047,1,1.08883)$, and the CIE cube-root transformation with a linear branch below 0.008856. HSV used the standard hexcone transformation. Hue was rotated by 0.15 modulo one to place red tissue within a contiguous histogram range rather than across the hue discontinuity.

Fixed channel ranges were chosen empirically to concentrate histogram resolution within the observed color distribution. Values were mapped into $[0,1)$ using
\begin{equation}
u_h=\operatorname{clip}\!\left(
\frac{v_h-\mathrm{lo}_h}{\mathrm{hi}_h-\mathrm{lo}_h},0,1-10^{-6}\right).
\end{equation}
The fixed endpoints and descriptor dimensions are given in~\autoref{tab:features}. These restricted ranges concentrated bins in the observed color distribution; theoretical chromatic ranges would have occupied only approximately three bins in the most informative channels.

\begin{table}[!htbp]
\caption{Color Normalization and Sector Descriptor Composition}
\label{tab:features}
\centering
\footnotesize
\setlength{\tabcolsep}{2bp}
\renewcommand{\arraystretch}{1.10}

\noindent\textbf{A. Channel scaling endpoints}\par\vspace{3bp}
\begin{tabular}{@{}p{0.34\columnwidth}p{0.27\columnwidth}p{0.27\columnwidth}@{}}
\toprule
Channel & Lower endpoint & Upper endpoint \\
\midrule
$L^*$ & 25 & 95 \\
$a^*$ & $-15$ & 40 \\
$b^*$ & $-15$ & 55 \\
Rotated hue & 0 & 0.45 \\
Saturation & 0 & 0.7 \\
Value & 0.3 & 1.0 \\
\bottomrule
\end{tabular}

\par\vspace{7bp}
\noindent\textbf{B. Descriptor blocks}\par\vspace{3bp}
\begin{tabular}{@{}p{0.39\columnwidth}p{0.15\columnwidth}p{0.38\columnwidth}@{}}
\toprule
Block & Dim. & Construction \\
\midrule
Univariate histograms & 96 & Six channels, 16 bins each \\
Joint $a^*$--$b^*$ histogram & 64 & $8\times8$ bins \\
Channel moments & 12 & Mean and SD for six channels \\
Glare fraction & 1 & Fraction of pixels classified as glare \\
\textbf{Own-sector descriptor} & \textbf{173} & Histograms, moments, and glare \\
Compact context & 65 & Five blocks of 13 moment/glare features \\
Position & 7 & Ring indicator, radial coordinate, sine, and cosine \\
\textbf{Emission input} & \textbf{245} & Own descriptor, context, and position \\
\bottomrule
\end{tabular}

\par\vspace{4bp}
\raggedright\footnotesize
Histograms were normalized independently. Moments and histograms excluded glare. Values outside the channel ranges were clamped after scaling.
\end{table}

Each sector descriptor contained six normalized 16-bin univariate histograms, one normalized $8\times8$ joint $a^*$--$b^*$ histogram, six means, six standard deviations, and the glare fraction, totaling 173 dimensions. For nonglare sector pixels $Q_i$, the moments were
\begin{equation}
\begin{aligned}
m_{i,h}&=|Q_i|^{-1}\sum_{p\in Q_i}u_h(p),\\
s_{i,h}&=\sqrt{\max\!\left(0,
 |Q_i|^{-1}\sum_{p\in Q_i}u_h(p)^2-m_{i,h}^2\right)}.
\end{aligned}
\end{equation}

\subsection{Neighbor Context and Positional Encoding}
The emission representation augmented the 173-dimensional descriptor $d_i$ with compact spatial context. Let $g_i\in\mathbb{R}^{13}$ contain the 12 moments and glare fraction, and let $\bar g=28^{-1}\sum_i g_i$. Five 13-dimensional blocks represented the inner neighbor, outer neighbor, average of the angular neighbors, frame mean, and the difference between the sector and frame mean:
\begin{equation}
\begin{aligned}
x_i=\big[&d_i,g_{\mathrm{in}(i)},g_{\mathrm{out}(i)},\\
&\tfrac12(g_{\mathrm{left}(i)}+g_{\mathrm{right}(i)}),
\bar g,g_i-\bar g,\pi_i\big].
\end{aligned}
\end{equation}
The seven-dimensional positional vector comprised a four-component ring indicator, normalized radial position, and sine and cosine of angular position:
\begin{equation}
\begin{aligned}
\pi_i&=[\operatorname{onehot}(k_i),(k_i+0.5)/4,
 \sin\phi_i,\cos\phi_i],\\
\phi_i&=2\pi a_i/M_{k_i}.
\end{aligned}
\end{equation}
The final representation had $173+5(13)+7=245$ dimensions. Feature ablations retained the same classifier and folds while replacing the compact context with 15 principal components per context block (255 dimensions in total), or omitting context and retaining only the own-sector descriptor and position (180 dimensions).

\subsection{Emission Classifier and Video-Grouped Validation}
The emission model used histogram-based gradient boosting with a maximum tree depth of six, minimum leaf size of 20, learning rate of 0.1, and up to 400 boosting iterations. Classifier settings were fixed across validation folds and feature ablations.

Training observations were weighted inversely to their class frequencies. For $N$ training sectors, $C$ classes, and class count $N_c$, the weight of observation $n$ was
\begin{equation}
\omega_n=\frac{N}{C N_{y_n}}.
\end{equation}
The resulting class-balanced posterior was treated as an approximation to a scaled class-conditional likelihood, $p_{\mathrm{bal}}(c\mid x)\propto p(x\mid c)$, for subsequent spatial and temporal inference. The emission classifier itself used no preceding or subsequent frames.

Validation design can influence assessment of surgical-video recognition models.\citep{Kostiuchik2024Surgical} We therefore used five-fold cross-validation grouped by recording, ensuring that all observations from a recording remained within a single fold. The same fold assignments were used for all compared models. For each held-out fold, the emission model was trained on the remaining recordings and produced nine-class posterior probabilities for the complete temporal grid of the held-out recordings.

\subsection{Learned Temporal Smoothing and Cross-Fitting}
Temporal stacking used a second gradient-boosting classifier to combine emission probabilities over short and longer windows. For sector $i$ at grid index $t$, define
\begin{equation}
\begin{aligned}
W_t[\ell,h]&=\{t+\ell,\ldots,t+h\}\cap\{1,\ldots,T\},\\
\bar p_t^{(i)}[\ell,h]&=\frac{1}{|W_t[\ell,h]|}
 \sum_{u\in W_t[\ell,h]}p_u^{(i)}.
\end{aligned}
\end{equation}
An empty window at a sequence boundary was replaced by the current posterior. The 52-dimensional smoother input was
\begin{equation}
\begin{aligned}
z_t^{(i)}=\big[&p_t^{(i)},\bar p_t^{(i)}[-6,-1],
 \bar p_t^{(i)}[-30,-7],\\
&\bar p_t^{(i)}[1,6],\bar p_t^{(i)}[7,30],\pi_i\big].
\end{aligned}
\end{equation}
On the regular 60-fps grid, these windows represented approximately the preceding and following one second and one-to-five seconds. Window lengths were fixed in advance in grid steps. A causal formulation retained only current and past posterior blocks, yielding 34 dimensions; the reported full pipeline used the bidirectional formulation. The smoother used the same boosting settings and class-balanced weights as the emission model and was trained at labeled keyframe sectors before prediction on the entire grid.

Training inputs for the temporal stage were obtained by nested cross-fitting within each outer training set. Emission predictions used to train the temporal model were generated out of fold, so no recording was represented by predictions from an emission model fitted on that same recording. The temporal model was then applied to emission predictions for the held-out outer fold. This design prevented second-stage training from benefiting from in-sample first-stage predictions.

\subsection{Within-Frame Spatial Regularization}
Spatial regularization used a Markov random field (MRF) over adjacent sectors. For each neighbor direction $\nu$, training-keyframe counts were collected where both sectors had labels, with add-one smoothing. Class coupling was defined by pointwise mutual information:
\begin{equation}
\Psi_\nu(c,c')=\log\frac{\widehat P_\nu(c,c')}
 {\widehat P_\nu(c)\widehat P_\nu(c')}.
\end{equation}
Let $r_i(c)$ denote the incoming posterior for one frame. It was the emission posterior in the spatial-only comparison and the learned smoother's output in the complete pipeline. Mean-field inference initialized $q_i=r_i$ and used the message
\begin{equation}
m_i(c)=\sum_{\nu:\nu(i)\ne i}\sum_{c'}
 q_{\nu(i)}(c')\Psi_\nu(c,c').
\end{equation}
Each iteration formed and damped an updated distribution:
\begin{equation}
\begin{aligned}
\widetilde q_i(c)&\propto\exp\{\log r_i(c)+\beta m_i(c)\},\\
q_i&\leftarrow\lambda q_i+(1-\lambda)\widetilde q_i.
\end{aligned}
\end{equation}
The coupling strength was $\beta=0.5$, damping was $\lambda=0.5$, and ten iterations were applied independently at each time-grid location. Pairwise counts were estimated from training folds only.

Two alternative spatial formulations were examined. A hierarchical model represented latent frame depth by the median labeled class and used the ratio $P(c\mid z,\mathrm{ring})/P(c)$ to relate sector classes to depth, with exact inference. A learned spatial-stacking model used 61 features derived from the sector's posterior, neighboring posteriors, and the frame mean. Its second-stage booster was trained using inner out-of-fold first-stage predictions. Additional comparisons combined spatial stacking with a banded HMM or applied cross-fitted per-class decision biases after the MRF.

\subsection{Anatomically Ordered Hidden Markov Models}
An independent nine-state HMM was applied to each sector's time sequence, with states ordered by tissue code. The banded transition matrix allowed persistence, one-state advancement, or one-state retreat:
\begin{equation}
\begin{aligned}
A_{c,c+1}&=\rho_c^{\uparrow},\qquad A_{c,c-1}=\rho_c^{\downarrow},\\
A_{c,c}&=1-\rho_c^{\uparrow}-\rho_c^{\downarrow},\\
A_{c,c'}&=0\quad\text{for }|c-c'|>1.
\end{aligned}
\end{equation}
The model prohibited skipping a state within a single grid step; multiple intervening steps could connect different tissues between sparsely annotated keyframes.

Transition rates were estimated separately for each training fold. For consecutive labeled keyframes of a sector with classes $a$ and $b$, separated by $n$ grid steps, an advance from $a$ to $b>a$ credited one exit to each state $a,\ldots,b-1$ and divided the $n$ dwell steps equally among $a,\ldots,b$. Retreats were handled analogously, crediting exits from $a,\ldots,b+1$. If $a=b$, all $n$ steps were assigned as dwell time in state $a$. For each permitted direction, the estimated rate was
\begin{equation}
\rho_c^{\uparrow}=\operatorname{clip}\!\left(
\frac{N_c^{\uparrow}}{D_c},10^{-3},0.5\right),
\end{equation}
with the analogous expression for retreat, where $D_c$ was credited dwell time and $N_c^{\uparrow}$ the credited advance count. The initial-state distribution used class frequencies at the first labeled keyframe of training recordings with add-one smoothing. In the strict left-to-right variant, retreat rates were zero and backward annotation changes were counted as dwelling.

The incoming posterior $r_t^{(i)}$ was converted to an emission score by
\begin{equation}
\log b_t^{(i)}(c)=\eta\log\max\{r_t^{(i)}(c),10^{-7}\}.
\end{equation}
The exponent $\eta$ was one in the adopted pipeline; values 0.1, 0.25, and 0.5 were evaluated to discount correlated evidence. In the full pipeline, $r_t^{(i)}$ was the spatially regularized posterior.

Forward--backward inference was performed in log space.\citep{Rabiner1989A} Writing $u_t(c)$ and $v_t(c)$ for log forward and backward messages and suppressing the sector index,
\begin{equation}
\begin{aligned}
u_t(c)&=\log b_t(c)\\
 &\quad+\operatorname{LSE}_{c'}[u_{t-1}(c')+\log A_{c',c}],\\
v_t(c)&=\operatorname{LSE}_{c'}[\log A_{c,c'}+\log b_{t+1}(c')\\
 &\hspace{80pt}+v_{t+1}(c')],
\end{aligned}
\end{equation}
where $\operatorname{LSE}$ denotes log-sum-exp. Smoothed state probabilities were
\begin{equation}
\gamma_t(c)=\frac{\exp\{u_t(c)+v_t(c)\}}
 {\sum_{c'}\exp\{u_t(c')+v_t(c')\}}.
\end{equation}
Log-space computation maintained numerical stability when forward and backward mass concentrated on different states. Viterbi decoding was additionally examined as a single-path alternative to posterior decoding.\citep{Rabiner1989A}

\subsection{Posterior Blending and Calibration Diagnostics}
The final sequence of operations was emission classification, learned temporal smoothing, MRF regularization, HMM smoothing, and geometric blending with the original emission posterior. The final probability was
\begin{equation}
\widehat p_t^{(i)}(c)=
\frac{\gamma_t^{(i)}(c)^{1-w}p_t^{(i)}(c)^w}
 {\sum_{c'}\gamma_t^{(i)}(c')^{1-w}p_t^{(i)}(c')^w},
\end{equation}
with $w=0.75$. Blending combined the raw classifier's class ranking with spatially and temporally regularized probabilities. Values $w\in\{0.25,0.5,0.75\}$ were examined in the pipeline without learned temporal smoothing. The selected blend weight and MRF coupling were carried over unchanged when the learned smoother was introduced.

Temperature scaling was evaluated separately as a log-loss diagnostic.\citep{Guo2017On} A scalar temperature $\mathcal{T}$ for each test fold was fitted using the other folds' out-of-fold outputs, with transformed probabilities proportional to $p(c)^{1/\mathcal{T}}$. This transformation preserves the maximum-probability class and therefore leaves accuracy and F1 unchanged. The diagnostic temperature results are reported separately from the unadjusted probability metrics in the main comparisons.

\subsection{Classification and Spatial-Consistency Endpoints}
The evaluation unit was a labeled sector on an annotated keyframe, scored using the outer-fold prediction for its video. Primary evaluation used clinician-only labels; secondary evaluation used all labels, including transferred targets. For fold $k$, let $\mathcal{C}_k$ be the classes present and $n_k$ the evaluated sector count. Accuracy, balanced accuracy, and macro-F1 were
\begin{equation}
\begin{aligned}
\mathrm{Acc}_k&=n_k^{-1}\sum_{n=1}^{n_k}\mathbf{1}\{\hat y_n=y_n\},\\
\mathrm{BAcc}_k&=|\mathcal{C}_k|^{-1}\sum_{c\in\mathcal{C}_k}
 \frac{\mathrm{TP}_c}{\mathrm{TP}_c+\mathrm{FN}_c},\\
\mathrm{F1}_k&=|\mathcal{C}_k|^{-1}\sum_{c\in\mathcal{C}_k}
 \frac{2\mathrm{TP}_c}{2\mathrm{TP}_c+\mathrm{FP}_c+\mathrm{FN}_c}.
\end{aligned}
\end{equation}
Macro AUROC was the mean one-versus-rest area under the receiver operating characteristic curve across $\mathcal{C}_k$. Log-loss was
\begin{equation}
\mathcal{L}_k=-n_k^{-1}\sum_{n=1}^{n_k}\log\widetilde p_n(y_n),
\end{equation}
where probabilities were floored at $10^{-7}$ and renormalized. Accuracy, macro-F1, and macro AUROC were the headline endpoints; balanced accuracy and log-loss characterized recall balance and probability quality.

Metrics were summarized as means and standard deviations across the five folds. Paired macro-F1 differences were calculated within each fold relative to the emission baseline, summarized by their mean and standard deviation, and accompanied by the number of positive folds. Per-class F1 was also computed from pooled held-out predictions. A training-fold class-frequency predictor served as a reference. Within-frame consistency was assessed on keyframes with at least eight clinician-annotated sectors. Depth span was the largest minus smallest tissue code on those sectors; the reported endpoint was the proportion of frames with span at least three, evaluated on the same sectors for annotations and predictions.

\subsection{Temporal Coherence and Peritoneal-Onset Endpoints}
Temporal endpoints used the entire temporal grid, including unlabeled frames. Flicker was defined as label changes per sector-minute:
\begin{equation}
\mathrm{Flicker}=\frac{\sum_{i=1}^{28}\sum_{t=2}^{T}
 \mathbf{1}\{\hat y_t^{(i)}\ne\hat y_{t-1}^{(i)}\}}
 {28(t_{\mathrm{end}}-t_{\mathrm{start}})/60}.
\end{equation}
The same calculation between consecutive clinician-labeled keyframes supplied a sparse-annotation reference. Backward share was the proportion of predicted label changes leading to a smaller tissue code. All elapsed-time denominators used the recording's own frame rate.

Peritoneum-or-deeper onset was evaluated for video--sector sequences whose clinician labels progressed from $y<7$ to $y\geq7$. The onset reference was the interval $[t_{\mathrm{lo}},t_{\mathrm{hi}}]$ bounded by the last shallower keyframe and first peritoneum-or-deeper keyframe. The predicted onset $\hat t$ was the first grid location assigned code at least seven. Error was distance outside the reference interval:
\begin{equation}
e=\max\{0,t_{\mathrm{lo}}-\hat t,\hat t-t_{\mathrm{hi}}\}.
\end{equation}
Reported measures were the proportion with $e\leq2$ seconds, premature onsets with $\hat t<t_{\mathrm{lo}}$, missed onsets when no qualifying prediction occurred, and median interval error. Predictions within the reference interval had zero error. These were retrospective sequence-recognition endpoints calculated from the evaluated model outputs.

\section{Results}
\subsection{Dataset Assembly and Preprocessing Verification}
The final analysis contained 71 videos comprising 21 densely annotated and 50 transferred recordings, with 1,516 keyframes containing at least one labeled sector, 36,886 all-label sectors, and 29,151 clinician-only sectors~(\autoref{tab:cohort}). Of the clinician-only sectors, 9,410 came from dense recordings and 19,741 from transferred recordings.

Temporal annotation alignment was verified before analysis. On dense reannotations, the fitted FoV contained 93\%--98\% of clinician-annotated pixels. Across clinician-only targets, rectus muscle was the most frequent class, with 7,876 sectors; bowel and preperitoneal fat had 503 and 537 sectors, respectively. All nine tissue classes were retained in the reported aggregate evaluation.

\subsection{Annotation-Transfer Validation}
The adopted transfer procedure, evaluated against dense annotations on recordings with both annotation paradigms, reconstructed dense clinician annotations with a mean per-class pixel IoU of 0.873 (class-specific range, 0.70--0.999). Extending filling to other historically plausible labels assigned the correct added label in 89\% of isolated assessments, but decreased overall IoU to 0.792. The subcutaneous-only transfer was therefore retained.

Across all training labels, 8,517 of 10,890 subcutaneous-fat sectors (approximately 78\%) originated from filling. Every subcutaneous-fat pixel in transferred recordings was automatically introduced. Accordingly, primary subcutaneous-fat evaluation was supplied entirely by densely annotated recordings. Filling could also change sector-majority labels when added fat pixels outnumbered a thin clinician-drawn layer; the all-label targets consequently contained fewer Scarpa's fascia sectors than the clinician-only targets.

\subsection{Primary Classification Results}
The frame-independent emission model achieved accuracy of $0.600\pm0.033$, macro-F1 of $0.440\pm0.045$, and macro AUROC of $0.861\pm0.024$ on clinician-annotated sectors~(\autoref{tab:primary}). The class-prior reference achieved accuracy of $0.084\pm0.024$, macro-F1 of $0.017\pm0.005$, and AUROC of 0.500. Learned temporal smoothing alone increased accuracy to $0.615\pm0.031$, macro-F1 to $0.469\pm0.036$, and AUROC to $0.879\pm0.018$. Its paired macro-F1 improvement was $0.029\pm0.020$, with improvement in four of five folds.

The complete pipeline achieved accuracy of $0.646\pm0.037$, macro-F1 of $0.507\pm0.049$, and AUROC of $0.882\pm0.018$, the highest values for all three primary classification endpoints among the compared methods. Relative to the emission baseline, the increases in reported means were 0.046 for accuracy and 0.021 for AUROC. The paired macro-F1 gain was $0.066\pm0.022$, with improvement in all five folds. Balanced accuracy increased from $0.452\pm0.042$ to $0.517\pm0.051$.

The pipeline without learned temporal smoothing, comprising MRF regularization, HMM smoothing, and blending, achieved accuracy of $0.628\pm0.040$, macro-F1 of $0.474\pm0.051$, and AUROC of $0.868\pm0.024$. Adding the learned smoother increased these reported means by 0.018, 0.033, and 0.014, respectively. Learned temporal smoothing alone accounted for most of the increase in AUROC relative to the emission model.

Spatial regularization alone produced macro-F1 of $0.462\pm0.044$, a paired improvement of $0.021\pm0.015$ in four folds. The strict HMM produced accuracy of $0.622\pm0.045$ but macro-F1 of $0.429\pm0.041$, yielding a paired F1 change of $-0.011\pm0.023$. The banded HMM produced macro-F1 of $0.443\pm0.048$, with a paired change of $0.003\pm0.019$. Combining the MRF and HMM produced macro-F1 of $0.462\pm0.049$ and a positive paired change in every fold. Adding the blend raised macro-F1 to 0.474 and AUROC from 0.827 to 0.868.

\begin{table*}[!htbp]
\caption{Primary Performance on Clinician-Annotated Sectors}
\label{tab:primary}
\centering\small\setlength\tabcolsep{3bp}\renewcommand\arraystretch{1.14}
\noindent\textbf{A. Classification and paired macro-F1 differences}\par\vspace{3bp}
\begin{tabular*}{\textwidth}{@{\extracolsep{\fill}}lcccc@{}}
\toprule
Model & Accuracy & Macro-F1 & Macro AUROC & $\Delta$F1 (positive folds) \\
\midrule
Class-prior reference & $0.084 \pm 0.024$ & $0.017 \pm 0.005$ & $0.500 \pm 0.000$ & $-0.423 \pm 0.046 (0/5)$ \\
Emission model & $0.600 \pm 0.033$ & $0.440 \pm 0.045$ & $0.861 \pm 0.024$ & \textemdash{} \\
Learned temporal smoothing & $0.615 \pm 0.031$ & $0.469 \pm 0.036$ & $0.879 \pm 0.018$ & $+0.029 \pm 0.020 (4/5)$ \\
MRF & $0.603 \pm 0.036$ & $0.462 \pm 0.044$ & $0.854 \pm 0.028$ & $+0.021 \pm 0.015 (4/5)$ \\
Strict left-to-right HMM & $0.622 \pm 0.045$ & $0.429 \pm 0.041$ & $0.816 \pm 0.014$ & $-0.011 \pm 0.023 (1/5)$ \\
Banded HMM & $0.625 \pm 0.044$ & $0.443 \pm 0.048$ & $0.838 \pm 0.045$ & $+0.003 \pm 0.019 (2/5)$ \\
MRF $\rightarrow$ HMM & $0.617 \pm 0.040$ & $0.462 \pm 0.049$ & $0.827 \pm 0.038$ & $+0.022 \pm 0.009 (5/5)$ \\
MRF $\rightarrow$ HMM $\rightarrow$ blend & $0.628 \pm 0.040$ & $0.474 \pm 0.051$ & $0.868 \pm 0.024$ & $+0.033 \pm 0.010 (5/5)$ \\
\textbf{Full pipeline} & {\bfseries\boldmath $0.646 \pm 0.037$} & {\bfseries\boldmath $0.507 \pm 0.049$} & {\bfseries\boldmath $0.882 \pm 0.018$} & {\bfseries\boldmath $+0.066 \pm 0.022 (5/5)$} \\
\bottomrule
\end{tabular*}
\par\vspace{8bp}
\noindent\textbf{B. Recall balance, log-loss, and spatial spread}\par\vspace{3bp}
\begin{tabular*}{\textwidth}{@{\extracolsep{\fill}}lccc@{}}
\toprule
Model & Balanced accuracy & Log-loss & Depth span $\geq3$ \\
\midrule
Class-prior reference & $0.111 \pm 0.000$ & $2.060 \pm 0.066$ & 0\% \\
Emission model & $0.452 \pm 0.042$ & $1.350 \pm 0.160$ & 53\% \\
Learned temporal smoothing & $0.483 \pm 0.037$ & {\bfseries\boldmath $1.233 \pm 0.157$} & 56\% \\
MRF & $0.473 \pm 0.041$ & $2.594 \pm 0.430$ & 21\% \\
Strict left-to-right HMM & $0.452 \pm 0.042$ & $4.797 \pm 0.578$ & 37\% \\
Banded HMM & $0.463 \pm 0.047$ & $4.003 \pm 0.638$ & 39\% \\
MRF $\rightarrow$ HMM & $0.472 \pm 0.048$ & $4.749 \pm 0.648$ & 22\% \\
MRF $\rightarrow$ HMM $\rightarrow$ blend & $0.482 \pm 0.050$ & $1.872 \pm 0.279$ & 25\% \\
\textbf{Full pipeline} & {\bfseries\boldmath $0.517 \pm 0.051$} & $1.693 \pm 0.266$ & 26\% \\
\bottomrule
\end{tabular*}
\par\vspace{4bp}\raggedright\footnotesize Values are means $\pm$ standard deviations across five video-grouped folds. $\Delta$F1 is the paired within-fold difference from emissions; parentheses give folds with a positive difference. The full pipeline is learned temporal smoothing $\rightarrow$ MRF $\rightarrow$ banded HMM $\rightarrow$ blend. MRF--HMM--blend omits learned smoothing. Depth span is measured on clinician-only sectors; the annotated reference is 22\%.
\end{table*}

\subsection{Secondary Evaluation Using All Labels}
When automatically transferred targets were included, the emission baseline achieved accuracy of $0.656\pm0.035$, macro-F1 of $0.450\pm0.045$, and AUROC of $0.867\pm0.031$~(\autoref{tab:all}). The full pipeline achieved $0.682\pm0.040$, $0.495\pm0.050$, and $0.886\pm0.024$, respectively. Its paired F1 improvement was $0.045\pm0.012$, with gains in every fold. Balanced accuracy increased from $0.452\pm0.045$ to $0.513\pm0.057$.

The full pipeline had the highest macro-F1, macro AUROC, and balanced accuracy in this evaluation. The strict HMM had the highest accuracy, $0.700\pm0.030$, followed by the banded HMM at $0.691\pm0.036$. Strict-HMM macro-F1 was $0.450\pm0.034$, with a paired change of $0.000\pm0.020$ and improvement in two folds. The class-prior reference achieved accuracy of $0.298\pm0.044$ and macro-F1 of $0.051\pm0.006$.

Learned temporal smoothing alone increased macro-F1 to $0.468\pm0.039$ and AUROC to $0.883\pm0.022$. The MRF--HMM--blend pipeline without the learned smoother achieved macro-F1 of $0.476\pm0.050$ and AUROC of $0.873\pm0.032$. Thus, adding the learned temporal stage increased the respective reported means by 0.019 and 0.013 under the all-label evaluation. The full set of probability, recall-balance, and spatial-spread measurements is retained in~\autoref{tab:all}.

\begin{table*}[!htbp]
\caption{Secondary Performance on All Labeled Sectors}
\label{tab:all}
\centering\small\setlength\tabcolsep{3bp}\renewcommand\arraystretch{1.14}
\noindent\textbf{A. Classification and paired macro-F1 differences}\par\vspace{3bp}
\begin{tabular*}{\textwidth}{@{\extracolsep{\fill}}lcccc@{}}
\toprule
Model & Accuracy & Macro-F1 & Macro AUROC & $\Delta$F1 (positive folds) \\
\midrule
Class-prior reference & $0.298 \pm 0.044$ & $0.051 \pm 0.006$ & $0.500 \pm 0.000$ & $-0.399 \pm 0.047 (0/5)$ \\
Emission model & $0.656 \pm 0.035$ & $0.450 \pm 0.045$ & $0.867 \pm 0.031$ & \textemdash{} \\
Learned temporal smoothing & $0.665 \pm 0.034$ & $0.468 \pm 0.039$ & $0.883 \pm 0.022$ & $+0.017 \pm 0.013 (4/5)$ \\
MRF & $0.641 \pm 0.040$ & $0.457 \pm 0.046$ & $0.859 \pm 0.035$ & $+0.007 \pm 0.014 (4/5)$ \\
Strict left-to-right HMM & {\bfseries\boldmath $0.700 \pm 0.030$} & $0.450 \pm 0.034$ & $0.827 \pm 0.015$ & $+0.000 \pm 0.020 (2/5)$ \\
Banded HMM & $0.691 \pm 0.036$ & $0.461 \pm 0.046$ & $0.852 \pm 0.048$ & $+0.010 \pm 0.014 (3/5)$ \\
MRF $\rightarrow$ HMM & $0.662 \pm 0.038$ & $0.460 \pm 0.049$ & $0.838 \pm 0.046$ & $+0.010 \pm 0.010 (4/5)$ \\
MRF $\rightarrow$ HMM $\rightarrow$ blend & $0.677 \pm 0.038$ & $0.476 \pm 0.050$ & $0.873 \pm 0.032$ & $+0.026 \pm 0.007 (5/5)$ \\
\textbf{Full pipeline} & $0.682 \pm 0.040$ & {\bfseries\boldmath $0.495 \pm 0.050$} & {\bfseries\boldmath $0.886 \pm 0.024$} & {\bfseries\boldmath $+0.045 \pm 0.012 (5/5)$} \\
\bottomrule
\end{tabular*}
\par\vspace{8bp}
\noindent\textbf{B. Recall balance, log-loss, and spatial spread}\par\vspace{3bp}
\begin{tabular*}{\textwidth}{@{\extracolsep{\fill}}lccc@{}}
\toprule
Model & Balanced accuracy & Log-loss & Depth span $\geq3$ \\
\midrule
Class-prior reference & $0.111 \pm 0.000$ & $1.840 \pm 0.059$ & 0\% \\
Emission model & $0.452 \pm 0.045$ & $1.142 \pm 0.160$ & 53\% \\
Learned temporal smoothing & $0.479 \pm 0.043$ & {\bfseries\boldmath $1.059 \pm 0.153$} & 56\% \\
MRF & $0.472 \pm 0.046$ & $2.343 \pm 0.464$ & 21\% \\
Strict left-to-right HMM & $0.451 \pm 0.039$ & $3.740 \pm 0.448$ & 37\% \\
Banded HMM & $0.461 \pm 0.049$ & $3.309 \pm 0.612$ & 39\% \\
MRF $\rightarrow$ HMM & $0.471 \pm 0.052$ & $4.166 \pm 0.668$ & 22\% \\
MRF $\rightarrow$ HMM $\rightarrow$ blend & $0.482 \pm 0.052$ & $1.592 \pm 0.269$ & 25\% \\
\textbf{Full pipeline} & {\bfseries\boldmath $0.513 \pm 0.057$} & $1.469 \pm 0.263$ & 26\% \\
\bottomrule
\end{tabular*}
\par\vspace{4bp}\raggedright\footnotesize Values are means $\pm$ standard deviations across five video-grouped folds. $\Delta$F1 is the paired within-fold difference from emissions; parentheses give folds with a positive difference. The full pipeline is learned temporal smoothing $\rightarrow$ MRF $\rightarrow$ banded HMM $\rightarrow$ blend. MRF--HMM--blend omits learned smoothing. Depth span is measured on clinician-only sectors; the annotated reference is 22\%.
\end{table*}

\subsection{Within-Frame Tissue-Depth Spread}
Among eligible keyframes, 22\% of clinician annotations had a depth span of at least three tissue codes. The corresponding proportion was 53\% for the emission baseline and 56\% after learned temporal smoothing alone. MRF regularization reduced the proportion to 21\%. Strict and banded HMMs without the spatial stage produced proportions of 37\% and 39\%, respectively. The MRF--HMM combination yielded 22\%, increasing to 25\% after blending and 26\% in the full pipeline. These comparisons used the same clinician-annotated sectors regardless of the label set used for classification scoring.

\begin{figure}[!tbp]
\centering
\includegraphics[width=\columnwidth]{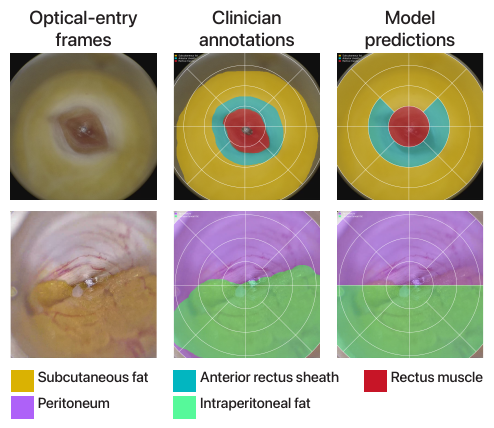}
\caption{Qualitative examples of spatially resolved tissue recognition. Columns show the optical-entry frame, clinician annotations, and model predictions on the polar grid. The upper row contains subcutaneous fat (yellow), anterior rectus sheath (cyan), and rectus muscle (red); the lower row contains peritoneum (purple) and intraperitoneal fat (green). Clinician-drawn boundaries vary within sectors, whereas predictions assign one tissue class to each sector. These examples illustrate the spatial representation and do not establish pixel-level segmentation accuracy.}
\label{fig:qualitative}
\end{figure}

\autoref{fig:qualitative} juxtaposes optical-entry frames, clinician annotations, and sector-level predictions. The examples illustrate the representation of multiple tissues within a frame and the coarser boundaries imposed by sector classification; they provide qualitative context for the aggregate spatial-consistency and classification results.

\subsection{Temporal Coherence and Onset Recognition}
The clinician annotations contained 1,774 observed video--sector onsets of peritoneum-or-deeper. The median width of the bracketing reference interval was 4.3 seconds. Between annotated keyframes, the reference label-change rate was $5.9\pm1.1$ per sector-minute and backward share was $0.148\pm0.039$~(\autoref{tab:temporal}). The emission model changed labels $66.2\pm3.7$ times per sector-minute. Learned temporal smoothing reduced this rate to $43.0\pm4.5$, whereas the spatial-only MRF yielded $61.2\pm3.0$.

The strict HMM produced the smallest label-change rate, $6.1\pm1.5$, with backward share $0.002\pm0.001$. The banded HMM produced $10.4\pm1.5$ changes per sector-minute and backward share $0.248\pm0.037$. The MRF--HMM combination yielded $19.2\pm2.6$ changes per sector-minute; blending yielded $20.6\pm1.4$; and the full pipeline yielded $19.5\pm3.5$. Relative to the emission model, the final rate was approximately 3.4-fold lower. Backward share decreased from $0.477\pm0.006$ to $0.382\pm0.022$ in the full pipeline.

The proportion of onsets within two seconds of the reference interval increased from $0.205\pm0.045$ for emissions to $0.344\pm0.051$ with learned temporal smoothing. The strict HMM achieved the largest proportion, $0.519\pm0.111$, followed by the banded HMM at $0.478\pm0.125$. The final pipeline achieved $0.466\pm0.099$. Premature-onset proportions were $0.828\pm0.044$ for the emission model, $0.669\pm0.058$ for learned smoothing, $0.225\pm0.083$ for the strict HMM, and $0.443\pm0.081$ for the complete pipeline.

Median interval error decreased from $15.58\pm5.03$ seconds for emissions to $6.95\pm2.73$ seconds after learned temporal smoothing. The strict HMM yielded $0.72\pm0.57$ seconds and the banded HMM $1.93\pm2.55$ seconds. The full pipeline yielded $2.45\pm1.29$ seconds, approximately 6.4-fold below the emission-model value. Missed-onset proportions were $0.012\pm0.019$ for emissions, $0.158\pm0.134$ for the strict HMM, and $0.162\pm0.104$ for the banded HMM. The complete pipeline yielded $0.054\pm0.080$, compared with $0.086\pm0.083$ for the MRF--HMM--blend pipeline without learned temporal smoothing.

\begin{table*}[!htbp]
\caption{Temporal Coherence and Peritoneum-or-Deeper Onset Recognition}
\label{tab:temporal}
\centering\small\setlength\tabcolsep{3bp}\renewcommand\arraystretch{1.14}
\noindent\textbf{A. Label changes and backward transitions}\par\vspace{3bp}
\begin{tabular*}{\textwidth}{@{\extracolsep{\fill}}lcc@{}}
\toprule
Model & Changes/sector-min & Backward share \\
\midrule
Clinician keyframes & $5.9 \pm 1.1$ & $0.148 \pm 0.039$ \\
Emission model & $66.2 \pm 3.7$ & $0.477 \pm 0.006$ \\
Learned temporal smoothing & $43.0 \pm 4.5$ & $0.452 \pm 0.010$ \\
MRF & $61.2 \pm 3.0$ & $0.472 \pm 0.007$ \\
Strict left-to-right HMM & {\bfseries\boldmath $6.1 \pm 1.5$} & $0.002 \pm 0.001$ \\
Banded HMM & $10.4 \pm 1.5$ & $0.248 \pm 0.037$ \\
MRF $\rightarrow$ HMM & $19.2 \pm 2.6$ & $0.339 \pm 0.026$ \\
MRF $\rightarrow$ HMM $\rightarrow$ blend & $20.6 \pm 1.4$ & $0.399 \pm 0.023$ \\
\textbf{Full pipeline} & $19.5 \pm 3.5$ & $0.382 \pm 0.022$ \\
\bottomrule
\end{tabular*}
\par\vspace{8bp}
\noindent\textbf{B. Onset timing relative to annotated intervals}\par\vspace{3bp}
\begin{tabular*}{\textwidth}{@{\extracolsep{\fill}}lcccc@{}}
\toprule
Model & Within 2 s & Premature & Missed & Median error (s) \\
\midrule
Clinician keyframes & \textemdash{} & \textemdash{} & \textemdash{} & \textemdash{} \\
Emission model & $0.205 \pm 0.045$ & $0.828 \pm 0.044$ & {\bfseries\boldmath $0.012 \pm 0.019$} & $15.58 \pm 5.03$ \\
Learned temporal smoothing & $0.344 \pm 0.051$ & $0.669 \pm 0.058$ & $0.027 \pm 0.049$ & $6.95 \pm 2.73$ \\
MRF & $0.263 \pm 0.081$ & $0.745 \pm 0.073$ & $0.017 \pm 0.023$ & $13.53 \pm 5.52$ \\
Strict left-to-right HMM & {\bfseries\boldmath $0.519 \pm 0.111$} & {\bfseries\boldmath $0.225 \pm 0.083$} & $0.158 \pm 0.134$ & {\bfseries\boldmath $0.72 \pm 0.57$} \\
Banded HMM & $0.478 \pm 0.125$ & $0.291 \pm 0.037$ & $0.162 \pm 0.104$ & $1.93 \pm 2.55$ \\
MRF $\rightarrow$ HMM & $0.459 \pm 0.085$ & $0.384 \pm 0.043$ & $0.097 \pm 0.089$ & $2.32 \pm 1.35$ \\
MRF $\rightarrow$ HMM $\rightarrow$ blend & $0.458 \pm 0.085$ & $0.423 \pm 0.055$ & $0.086 \pm 0.083$ & $2.71 \pm 1.45$ \\
\textbf{Full pipeline} & $0.466 \pm 0.099$ & $0.443 \pm 0.081$ & $0.054 \pm 0.080$ & $2.45 \pm 1.29$ \\
\bottomrule
\end{tabular*}
\par\vspace{4bp}\raggedright\footnotesize Values are means $\pm$ standard deviations across folds, including the fold summaries of median error. Onset frequencies are proportions. Error is distance outside the annotated interval; within-interval predictions have zero error. The clinician keyframe change rate is a sparse-sampling lower-bound reference. There were 1,774 observed sector onsets, with median reference-interval width 4.3 s.
\end{table*}

\subsection{Class-Specific Recognition}
\begin{figure}[!htbp]
\centering
\includegraphics[width=\columnwidth]{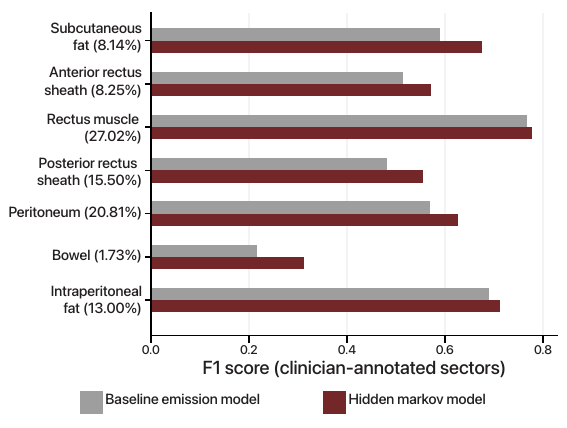}
\caption{Class-specific recognition for seven of the nine tissue classes. Gray bars show the emission baseline. Red bars, labeled ``Hidden markov model,'' represent the complete pipeline: learned temporal smoothing, MRF regularization, banded HMM inference, and blending. Their values correspond to the full-pipeline column of~\autoref{tab:classes}. Percentages give each class's share of the 29,151 clinician-annotated sectors. F1 values pool held-out predictions across folds. Scarpa's fascia and preperitoneal fat are omitted from this plot and retained in the table.}
\label{fig:class-f1}
\end{figure}

\autoref{fig:class-f1} compares the emission baseline with the full pipeline for seven tissue classes, using held-out predictions pooled across folds. The class frequencies shown alongside the bars highlight the imbalance in the clinician-annotated evaluation set; pooled held-out F1 values are reported for all nine classes in~\autoref{tab:classes}. Subcutaneous-fat F1 increased from 0.590 for the emission baseline to 0.676 for the full pipeline. Scarpa's fascia increased from 0.229 to 0.455, with an intermediate value of 0.381 after learned temporal smoothing alone. Anterior rectus sheath increased from 0.515 to 0.572, posterior rectus sheath from 0.481 to 0.555, and rectus muscle from 0.768 to 0.778.

Peritoneum F1 increased from 0.569 to 0.626, and bowel from 0.217 to 0.312. Intraperitoneal-fat F1 increased from 0.691 to 0.713; the largest value for that class was 0.720 with MRF--HMM--blend without learned smoothing. Preperitoneal-fat F1 values were 0.018 for emissions, 0.015 for learned smoothing, 0.025 for MRF--HMM--blend, and 0.024 for the full pipeline. Rectus-muscle F1 was 0.778 with either blended pipeline. These pooled class-specific estimates are distinct from fold-averaged macro-F1.

\begin{table*}[!htbp]
\caption{Pooled Class-Specific F1 on Clinician-Annotated Sectors}
\label{tab:classes}
\centering\small\setlength\tabcolsep{3bp}\renewcommand\arraystretch{1.14}
\begin{tabular*}{\textwidth}{@{\extracolsep{\fill}}lrcccc@{}}
\toprule
Tissue & Sectors & Emission & Temporal smoother & MRF--HMM--blend & Full pipeline \\
\midrule
Subcutaneous fat & 2,373 & 0.590 & 0.646 & 0.637 & {\bfseries\boldmath 0.676} \\
Scarpa's fascia & 1,082 & 0.229 & 0.381 & 0.313 & {\bfseries\boldmath 0.455} \\
Anterior sheath & 2,406 & 0.515 & 0.519 & 0.543 & {\bfseries\boldmath 0.572} \\
Rectus muscle & 7,876 & 0.768 & 0.756 & {\bfseries\boldmath 0.778} & {\bfseries\boldmath 0.778} \\
Posterior sheath & 4,517 & 0.481 & 0.501 & 0.525 & {\bfseries\boldmath 0.555} \\
Preperitoneal fat & 537 & 0.018 & 0.015 & {\bfseries\boldmath 0.025} & 0.024 \\
Peritoneum & 6,066 & 0.569 & 0.594 & 0.610 & {\bfseries\boldmath 0.626} \\
Bowel & 503 & 0.217 & 0.239 & 0.243 & {\bfseries\boldmath 0.312} \\
Intraperitoneal fat & 3,791 & 0.691 & 0.697 & {\bfseries\boldmath 0.720} & 0.713 \\
\bottomrule
\end{tabular*}
\par\vspace{4bp}\raggedright\footnotesize Values were computed from held-out predictions pooled across folds. MRF--HMM--blend omits learned temporal smoothing. Subcutaneous-fat primary targets came from dense recordings. Bold values identify the best F1 in each row, including ties.
\end{table*}

\subsection{Feature-Context Ablations}
The 180-dimensional own-descriptor-and-position model achieved clinician-only accuracy of $0.514\pm0.033$, macro-F1 of $0.402\pm0.032$, and AUROC of $0.833\pm0.022$~(\autoref{tab:ablation}). The 255-dimensional PCA-context representation yielded $0.578\pm0.027$, $0.428\pm0.038$, and $0.847\pm0.029$, respectively. The adopted 245-dimensional moments-and-glare representation yielded $0.600\pm0.033$, $0.440\pm0.045$, and $0.861\pm0.024$.

Relative to the adopted representation, omitting context reduced macro-F1 by approximately 0.038 and improved no folds. PCA context reduced macro-F1 by approximately 0.012 and improved two folds.

The same ordering was observed with all-label targets: macro-F1 was $0.398\pm0.036$ without context, $0.434\pm0.032$ with PCA context, and $0.450\pm0.045$ with compact context. Corresponding accuracies were 0.549, 0.635, and 0.656, and AUROCs were 0.835, 0.853, and 0.867. \autoref{tab:ablation} includes the reported classification and probability metrics for both label sets.

\begin{table*}[!htbp]
\caption{Complete Feature-Context Ablation Results}
\label{tab:ablation}
\centering\small\setlength\tabcolsep{3bp}\renewcommand\arraystretch{1.14}
\noindent\textbf{A. Clinician-only classification}\par\vspace{3bp}
\begin{tabular*}{\textwidth}{@{\extracolsep{\fill}}lcccc@{}}
\toprule
Features & Dimensions & Accuracy & Macro-F1 & Macro AUROC \\
\midrule
No context & 180 & $0.514 \pm 0.033$ & $0.402 \pm 0.032$ & $0.833 \pm 0.022$ \\
PCA-15 context & 255 & $0.578 \pm 0.027$ & $0.428 \pm 0.038$ & $0.847 \pm 0.029$ \\
Moments + glare & {\bfseries\boldmath 245} & {\bfseries\boldmath $0.600 \pm 0.033$} & {\bfseries\boldmath $0.440 \pm 0.045$} & {\bfseries\boldmath $0.861 \pm 0.024$} \\
\bottomrule
\end{tabular*}
\par\vspace{8bp}
\noindent\textbf{B. All-label classification}\par\vspace{3bp}
\begin{tabular*}{\textwidth}{@{\extracolsep{\fill}}lcccc@{}}
\toprule
Features & Dimensions & Accuracy & Macro-F1 & Macro AUROC \\
\midrule
No context & 180 & $0.549 \pm 0.042$ & $0.398 \pm 0.036$ & $0.835 \pm 0.027$ \\
PCA-15 context & 255 & $0.635 \pm 0.029$ & $0.434 \pm 0.032$ & $0.853 \pm 0.035$ \\
Moments + glare & {\bfseries\boldmath 245} & {\bfseries\boldmath $0.656 \pm 0.035$} & {\bfseries\boldmath $0.450 \pm 0.045$} & {\bfseries\boldmath $0.867 \pm 0.031$} \\
\bottomrule
\end{tabular*}
\par\vspace{8bp}
\noindent\textbf{C. Diagnostic metrics for both label sets}\par\vspace{3bp}
\begin{tabular*}{\textwidth}{@{\extracolsep{\fill}}llcc@{}}
\toprule
Features & Label set & Balanced accuracy & Log-loss \\
\midrule
No context & Clinician only & $0.436 \pm 0.029$ & $1.406 \pm 0.078$ \\
PCA-15 context & Clinician only & $0.441 \pm 0.030$ & $1.394 \pm 0.102$ \\
Moments + glare & Clinician only & {\bfseries\boldmath $0.452 \pm 0.042$} & {\bfseries\boldmath $1.350 \pm 0.160$} \\
No context & All labels & $0.436 \pm 0.033$ & $1.316 \pm 0.108$ \\
PCA-15 context & All labels & $0.438 \pm 0.032$ & $1.198 \pm 0.118$ \\
Moments + glare & All labels & {\bfseries\boldmath $0.452 \pm 0.045$} & {\bfseries\boldmath $1.142 \pm 0.160$} \\
\bottomrule
\end{tabular*}
\par\vspace{4bp}\raggedright\footnotesize Values are means $\pm$ standard deviations unless otherwise specified. All variants used identical model settings and validation folds. No-context inputs contained own-sector descriptors and position; PCA used 15 components per context block; the adopted representation used moments and glare.
\end{table*}

\subsection{Alternative Spatial Models and Decision Biases}
The hierarchical frame-depth prior with full emission weighting achieved clinician-only accuracy of $0.576\pm0.039$, macro-F1 of $0.439\pm0.045$, and AUROC of $0.854\pm0.020$~(\autoref{tab:alternatives}). Its macro-F1 was approximately unchanged from the emission baseline. Reducing evidence weighting to 0.5 or below produced collapse toward rare classes favored by the prior ratio. Learned spatial stacking achieved macro-F1 of $0.415\pm0.036$, approximately 0.025 below emissions, and improved no folds. Combining spatial stacking with the banded HMM yielded macro-F1 of $0.431\pm0.036$ and AUROC of $0.836\pm0.038$.

Cross-fitted per-class decision biases after the MRF yielded clinician-only accuracy of $0.602\pm0.040$, macro-F1 of $0.471\pm0.044$, and AUROC of $0.859\pm0.028$. Under all-label scoring, the corresponding reported means were 0.608, 0.445, and 0.853, below MRF-only results of 0.641, 0.457, and 0.859. Estimated biases reached the positive or negative three-unit search boundaries.

\begin{table*}[!htbp]
\caption{Alternative Spatial Models, Decision Biases, and Emission Weighting}
\label{tab:alternatives}
\centering\small\setlength\tabcolsep{3bp}\renewcommand\arraystretch{1.14}
\begin{tabular*}{\textwidth}{@{\extracolsep{\fill}}lccc@{}}
\toprule
Alternative & Accuracy & Macro-F1 & Macro AUROC \\
\midrule
Hierarchical frame-depth prior ($\alpha=1$) & $0.576 \pm 0.039$ & $0.439 \pm 0.045$ & $0.854 \pm 0.020$ \\
Learned spatial stacking & $0.571 \pm 0.040$ & $0.415 \pm 0.036$ & $0.838 \pm 0.017$ \\
Spatial stacking + banded HMM & $0.613 \pm 0.041$ & $0.431 \pm 0.036$ & $0.836 \pm 0.038$ \\
MRF + per-class decision bias & $0.602 \pm 0.040$ & $0.471 \pm 0.044$ & $0.859 \pm 0.028$ \\
HMM emission exponent $\eta<1$ & \textemdash{} & \textemdash{} & \textemdash{} \\
\bottomrule
\end{tabular*}
\par\vspace{8bp}
\begin{tabular*}{\textwidth}{@{\extracolsep{\fill}}>{\raggedright\arraybackslash}p{175bp}>{\raggedright\arraybackslash}p{270bp}@{}}
\toprule
Alternative & Experimental finding \\
\midrule
Hierarchical frame-depth prior ($\alpha=1$) & Approximately neutral at $\alpha=1$; collapsed when $\alpha\leq0.5$. \\
Learned spatial stacking & Below emissions; 0/5 folds improved. \\
Spatial stacking + banded HMM & Below MRF--HMM macro-F1. \\
MRF + per-class decision bias & All-label accuracy/F1/AUROC: 0.608/0.445/0.853, below MRF alone. Biases reached $\pm3$. \\
HMM emission exponent $\eta<1$ & Lower macro-F1 for every tested $\eta\in\{0.1,0.25,0.5\}$. \\
\bottomrule
\end{tabular*}
\par\vspace{4bp}\raggedright\footnotesize Metrics are clinician-only means $\pm$ standard deviations. The all-label decision-bias values are reported means. Dashes indicate metrics not numerically specified for the exponent sweep.
\end{table*}

\subsection{Parameter Sensitivity, Decoding, and Probability Diagnostics}
When the MRF operated directly on emission posteriors, macro-F1 improvement was retained over coupling strengths from 0.25 to 1. In the pipeline without learned temporal smoothing, each examined blend weight, 0.25, 0.5, and 0.75, improved accuracy, macro-F1, and AUROC relative to the emission model. Reducing the HMM emission exponent to 0.1, 0.25, or 0.5 did not improve macro-F1. Viterbi and forward--backward decoding differed by no more than 0.01 on the reported label metrics. These additional comparisons are consolidated in~\autoref{tab:diagnostics}.

Learned temporal smoothing alone produced the smallest clinician-only log-loss, $1.233\pm0.157$, compared with $1.350\pm0.160$ for emissions. Unblended strict and banded HMMs yielded $4.797\pm0.578$ and $4.003\pm0.638$, respectively. The MRF--HMM combination yielded $4.749\pm0.648$, decreasing to $1.872\pm0.279$ after blending and $1.693\pm0.266$ in the full pipeline. The all-label evaluation similarly favored learned smoothing for log-loss, with $1.059\pm0.153$ compared with $1.142\pm0.160$ for emissions and $1.469\pm0.263$ for the complete pipeline.

In the separate temperature-scaling analysis, the emission model required a temperature of approximately 1.5, the unblended banded HMM approximately 11, and the strict HMM approximately 70. The larger HMM temperatures indicated a need to substantially soften those posteriors for the diagnostic log-loss analysis. Temperature scaling did not change the predicted classes, accuracy, or F1, as expected for a positive scalar transformation that preserves the highest-probability class.\citep{Guo2017On}

\begin{table*}[!htbp]
\caption{Additional Ablations, Parameter Sweeps, and Calibration Diagnostics}
\label{tab:diagnostics}
\centering\small\setlength\tabcolsep{3bp}\renewcommand\arraystretch{1.14}
\begin{tabular*}{\textwidth}{@{\extracolsep{\fill}}>{\raggedright\arraybackslash}p{135bp}>{\raggedright\arraybackslash}p{130bp}>{\raggedright\arraybackslash}p{181bp}@{}}
\toprule
Experiment & Setting or comparison & Reported outcome \\
\midrule
MRF coupling & $\beta\in[0.25,1]$ & Macro-F1 gain over emissions retained \\
Geometric blend without temporal stacking & $w=0.25,0.5,0.75$ & All tested weights improved accuracy, macro-F1, and AUROC over emissions \\
HMM emission discounting & $\eta=0.1,0.25,0.5$ & No tested value improved macro-F1 \\
Posterior vs path decoding & Forward--backward vs Viterbi & Differences within $\pm0.01$ on label metrics \\
Hierarchical evidence weighting & $\alpha\leq0.5$ & Collapsed toward rare classes \\
PCA context vs adopted context & 255 vs 245 dimensions & Macro-F1 change $-0.012$; 2/5 folds improved \\
No context vs adopted context & 180 vs 245 dimensions & Macro-F1 change $-0.038$; 0/5 folds improved \\
Emission temperature scaling & $\mathcal{T}\approx1.5$ & Diagnostic probability softening \\
Banded-HMM temperature scaling & $\mathcal{T}\approx11$ & Diagnostic probability softening \\
Strict-HMM temperature scaling & $\mathcal{T}\approx70$ & Diagnostic probability softening \\
\bottomrule
\end{tabular*}
\par\vspace{4bp}\raggedright\footnotesize Qualitative sweep outcomes are reproduced as reported; unreported per-setting metrics are not imputed. Context-ablation fold counts refer to clinician-only evaluation. Temperature scaling preserves predicted classes, accuracy, and F1.
\end{table*}

\section{Discussion}
\subsection{Principal Findings}
In this retrospective study of 71 optical trocar recordings, integrating appearance, spatial context, and temporal structure improved tissue recognition across nine anatomical classes. On clinician-annotated sectors, the complete pipeline achieved accuracy of 0.646, macro-F1 of 0.507, and macro AUROC of 0.882, compared with 0.600, 0.440, and 0.861 for independent frame classification. Macro-F1 improved in all five folds. These gains accompanied fewer prediction changes and more accurate localization of peritoneum-or-deeper onset: flicker decreased from 66.2 to 19.5 changes per sector-minute, and median onset error decreased from 15.58 to 2.45 seconds. However, improved aggregate discrimination did not translate uniformly across tissues or endpoints.

\subsection{Context and Temporal Structure}
The ablations suggest distinct roles for each component. Neighbor context improved appearance classification, while learned temporal smoothing contributed most of the AUROC gain and improved recognition of Scarpa's fascia. Spatial regularization reduced anatomically disparate predictions within individual frames toward the distribution observed in clinician annotations. Temporal constraints suppressed rapid transitions, but stronger constraints introduced a tradeoff: the strict hidden Markov model achieved the lowest onset error while missing approximately 16\% of observed onsets. Blending preserved more appearance evidence, reducing missed onsets to approximately 5\% in the complete pipeline. Its premature-onset proportion nevertheless remained 44\%, underscoring the distinction between improved retrospective sequence labeling and dependable event detection. The study does not establish whether the observed improvements are sufficient for clinical use. In particular, greater agreement with anatomical ordering cannot substitute for correct identification of individual tissues, and visually smoother predictions may still conceal errors at clinically consequential transition points.

\subsection{Methodological Strengths}
The framework represents multiple tissues within a frame rather than imposing a single tissue label. Video-grouped evaluation separates recordings between training and testing,\citep{Kostiuchik2024Surgical} and nested cross-fitting prevents the temporal classifier from being trained on in-sample emission predictions from the same recordings. Primary targets were recomputed exclusively from clinician-drawn pixels, reducing dependence of evaluation on automatically transferred labels.\citep{Ward2021Challenges,Eckhoff2023SAGES} Common folds, component ablations, unsuccessful alternatives, and complementary classification, spatial, and temporal endpoints make the contributions and tradeoffs inspectable. The compact color descriptors and explicit anatomical constraints also provide an interpretable computational baseline for subsequent representation-learning studies.

\subsection{Limitations and Future Work}
The retrospective sample remains modest, and numerous sectors within each recording are correlated observations rather than independent cases. External validation was not reported; repeated model comparisons on the same folds may also produce optimistic selection. Annotation transfer expanded training coverage but introduced heuristic labels, while sparse keyframes limited temporal precision: the median reference interval for onset was 4.3 seconds. Preperitoneal-fat recognition remained poor, with F1 of 0.024, and bowel recognition remained limited at 0.312. Color features and fixed sectors may inadequately represent subtle boundaries, texture, or changing tissue geometry. Probability quality also requires attention.\citep{Guo2017On,Mehrtash2020Confidence,Kompa2021Second} The complete pipeline had higher log-loss than learned smoothing alone despite superior classification metrics. Finally, future temporal windows and forward--backward inference make the reported pipeline noncausal; these results do not establish real-time guidance performance. Prospective evaluation should separately assess causal prediction, calibration, external generalizability, and effects on surgeon decisions and patient outcomes,\citep{Hashimoto2018Artificial,Mascagni2022Artificial} with particular attention to premature alerts and missed entry events.

\bibliographystyle{vancouver}
\bibliography{references}
\end{document}